\documentclass[twocolumn,9pt]{article} 

\usepackage[square,numbers,sort&compress,comma]{natbib}

\usepackage{amsmath}
\usepackage{amssymb}
\usepackage{caption}
\usepackage{graphicx}
\usepackage{latexsym}
\usepackage{times}
\usepackage[pagewise]{lineno}
\usepackage{hyperref}

\topmargin - 12pt 
\renewenvironment{abstract}%
              {
               \small
               {\bfseries \abstractname}
               \par
               \vspace{10pt}
              }

\renewcommand\abstractname{Abstract}

\newcommand{\nomenclature}
              [1]
              {
               \bgroup
               \flushleft
               \small\bf
               #1
               \par
               \egroup
              }

\renewcommand{\section}
              [1]
              {
               \bgroup
               \flushleft
               \small\bf
               \refstepcounter{section}
               \arabic{section}. #1
               \par
               \egroup
              }

\renewcommand{\subsection}
              [1]
              {
               \bgroup
               \flushleft
               \small\em
               \refstepcounter{subsection}
               \arabic{section}.
               \arabic{subsection}. #1
               \par
               \egroup
              }

\renewcommand{\subsubsection}
              [1]
              {
               \bgroup
               \flushleft
               \small\em
               \refstepcounter{subsubsection}
               \arabic{section}.
               \arabic{subsection}.
               \arabic{subsubsection}. #1
               \par
               \egroup
              }

  \newcommand{\acknowledgement}
              [1]
              {
               \bgroup
               \flushleft
               \small\bf
               #1
               \par
               \egroup
              }

  \newcommand{\sectionbib}
              [1]
              {
               \bgroup
               \flushleft
               \small\bf
               #1
               \par
               \egroup
              }

\begin{document}



\small
\baselineskip 10pt

\setcounter{page}{1}
\title{\LARGE \bf Towards LLM-enabled autonomous\\
    combustion research: A literature-aware agent\\
    for self-corrective modeling workflows}

\author{{\large Ke Xiao$^{a,b}$, Haoze Zhang$^{a}$, Runze Mao$^{a}$, Han Li$^{a,b,*}$, Zhi X. Chen$^{a,b,*}$}\\[10pt]
        {\footnotesize \em $^a$State Key Laboratory of Turbulence and Complex Systems, School of Mechanics and Engineering Science,}\\[-5pt]
        {\footnotesize \em Peking University, Beijing 100871, China}\\[-5pt]
        {\footnotesize \em $^b$AI for Science Institute (AISI), Beijing 100080, China}}

\date{}  

\twocolumn[\begin{@twocolumnfalse}
\maketitle
\rule{\textwidth}{0.5pt}
\vspace{-5pt}

\begin{abstract} 
The rapid evolution of large language models (LLMs) is transforming artificial intelligence into autonomous research partners, yet a critical gap persists in complex scientific domains such as combustion modeling. Here, practical AI assistance requires the seamless integration of domain literature knowledge with robust execution capabilities for expertise-intensive tools such as computational fluid dynamics (CFD) codes. To bridge this gap, we introduce FlamePilot, an LLM agent designed to empower combustion modeling research through automated and self-corrective CFD workflows. FlamePilot differentiates itself through an architecture that leverages atomic tools to ensure the robust setup and execution of complex simulations in both OpenFOAM and extended frameworks such as DeepFlame. The system is also capable of learning from scientific articles, extracting key information to guide the simulation from initial setup to optimized results. Validation on a public benchmark shows FlamePilot achieved a perfect 1.0 executability score and a 0.438 success rate, surpassing the prior best reported agent scores of 0.625 and 0.250, respectively. Furthermore, a detailed case study on Moderate or Intense Low-oxygen Dilution (MILD) combustion simulation demonstrates its efficacy as a collaborative research copilot, where FlamePilot autonomously translated a research paper into a configured simulation, conducted the simulation, post-processed the results, proposed evidence-based refinements, and managed a multi-step parameter study to convergence under minimal human intervention. By adopting a transparent and interpretable paradigm, FlamePilot establishes a foundational framework for AI-empowered combustion modeling, fostering a collaborative partnership where the agent manages workflow orchestration, freeing the researcher for high-level analysis.
\end{abstract}

\vspace{5pt}
\parbox{1.0\textwidth}{\footnotesize {\em Keywords:} LLM agents; LLMs for research; Combustion modeling; CFD automation}
\rule{\textwidth}{0.5pt}
*Corresponding author.
\vspace{5pt}
\end{@twocolumnfalse}] 

\section{Introduction\label{sec:introduction}} \addvspace{10pt}
The rapid advancements in large language models (LLMs) are significantly influencing scientific research, transforming artificial intelligence (AI) from a computational aid to an autonomous research partner \cite{wei_ai_2025,wang_hitchhikers_2025}. By automating cognitive tasks such as hypothesis generation, experimental design, execution, and analysis, LLM-driven AI agent systems accelerate research cycles and enable researchers to dedicate more effort to creative and high-level thinking.

\begin{figure*}[ht!]
\centering
\vspace{-0.6 in}
\includegraphics[width=360pt]{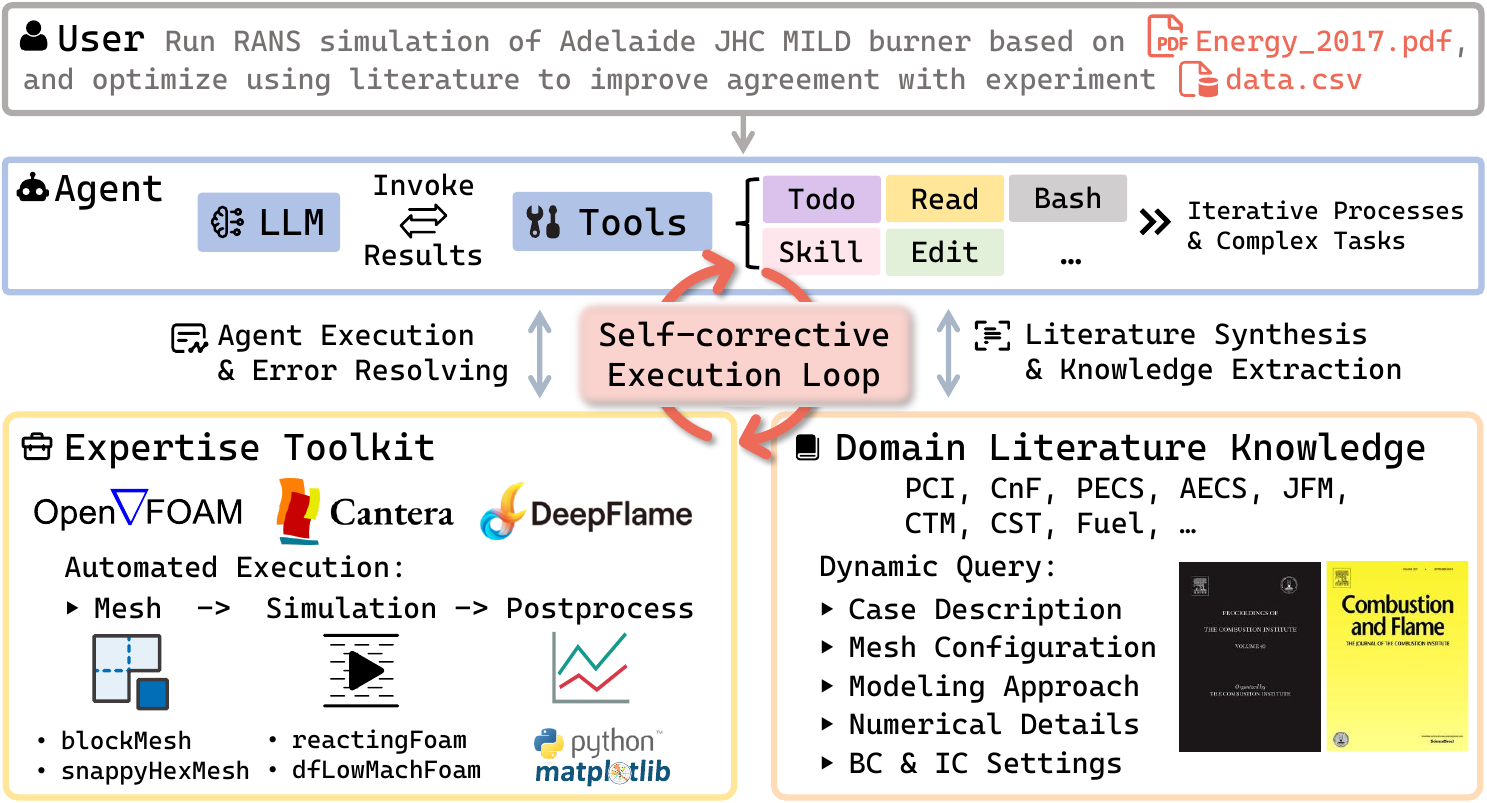}
\vspace{5 pt}
\caption{\footnotesize Schematic overview of the FlamePilot agent architecture, illustrating the integration of LLM reasoning, tool-based execution, and dynamic domain knowledge comprehension for automated CFD workflows.}
\label{overview}
\end{figure*}

Although significant progress has been made in developing such agents for both general research contexts \cite{chai_scimaster_2025,mitchener_kosmos_2025} and specialized fields such as mathematics \cite{novikov_alphaevolve_2025}, chemistry \cite{pham_chemgraph_2025}, and life sciences \cite{huang_biomni_2025}, the combustion research community notably still lacks a practical and accessible AI agent system. Such a system must integrate domain-specific knowledge with robust execution capabilities to effectively handle expertise-intensive tools such as OpenFOAM \cite{jasak_openfoam_2009} for computational fluid dynamics (CFD) applications and Cantera \cite{cantera} for chemical kinetics. 

While existing frameworks such as MetaOpenFOAM \cite{chen_metaopenfoam_2024,chen_optmetaopenfoam_2025,chen_metaopenfoam_2025} and others \cite{xu_llm_2024,dong_fine-tuning_2025,fan_chatcfd_2025,pandey_openfoamgpt_2025,feng_openfoamgpt_2025,xu_cfdagent_2025,yue_foam-agent_2025} have advanced OpenFOAM workflow automation, they remain insufficient for broad adoption. Benchmarking studies reveal low success rates on complex problems \cite{somasekharan_cfdllmbench_2025}, indicating a lack of reliable execution. Their practicality is further limited by a reliance on specific OpenFOAM versions, hindering compatibility with custom research solvers. Crucially, most of current agents cannot autonomously integrate knowledge from scientific literature, instead depending on user prompts to inject critical domain context. This inability to leverage foundational knowledge sources prevents their effective use in real research workflows.

This study develops FlamePilot, an LLM-powered research assistant for combustion modeling that integrates domain knowledge with robust OpenFOAM execution. The system operates as a human-supervised partner, performing self-corrective actions using execution feedback and scientific literature to ensure physically accurate and reproducible results. This architecture prioritizes verifiability and interpretability to establish the scientific rigor required for trustworthy AI assistance. Inspired by coding agents, FlamePilot handles OpenFOAM workflows through file operations and command-line execution, supporting both standard and extended frameworks like DeepFlame \cite{mao_deepflame_2023,mao_integrated_2024,mao_deepflame_2025}. Empirical validation on a Moderate or Intense Low-oxygen Dilution (MILD) combustion case confirms its efficacy as a research copilot, maintaining verifiability and interpretability throughout automated CFD workflows. This emphasis on verifiability and interpretability establishes the scientific rigor necessary for building trust on AI assistants and achieving reproducible outcomes in computational science.

\section{Agent system and architecture\label{sec:system-architecture}} \addvspace{10pt}
FlamePilot adapts the proven paradigm of command-line interface (CLI) coding agents \cite{noauthor_claude_2025,noauthor_opencode_2025,noauthor_gemini_2025,noauthor_codex_2025} to OpenFOAM workflows, enabling seamless integration as a "research copilot." As Fig. \ref{overview} illustrates, its architecture integrates three core components: a central LLM for workflow orchestration, an expertise toolkit for simulation control, and a literature synthesis module for domain knowledge. This unified design supports both automated execution and knowledge-informed assistance.

Departing from complex multi-agent systems, FlamePilot's pragmatic single-agent architecture prioritizes executability and transparency. Its use of atomic tools and specialized skills enables autonomous error correction while maintaining full researcher oversight. By keeping all decisions accessible and grounding simulations in dynamically extracted literature, this approach ensures the scientific rigor and reproducibility that opaque automation lacks.

\subsection{Tool Integration\label{subsec:tool-integration}} \addvspace{10pt}
Effective LLM agent operation in scientific computing requires versatile tool integration, a need unmet by existing CFD agents constrained by specialized, monolithic functions. FlamePilot addresses this through a foundation of atomic tools following established coding agent paradigms, including file read/write, directory listing, grep search, and bash execution. This granular toolset enables dynamic composition for complex CFD tasks: configuring dictionaries, executing solvers, and diagnosing errors via log analysis.

For knowledge retrieval from OpenFOAM tutorial cases, FlamePilot employs direct filesystem search over static RAG, leveraging OpenFOAM's key-value pair structure for efficient pattern matching in tutorial directories. Workflow orchestration is handled by a task management system providing decomposition, planning, and progress tracking, enabling autonomous management of multi-stage simulations. This reasoning mechanism, inspired by advanced coding agents \cite{noauthor_claude_2025,noauthor_opencode_2025}, allows FlamePilot to autonomously manage intricate, multi-stage simulations while preserving a transparent and auditable execution path for essential researcher oversight.

\subsection{Domain Knowledge Integration\label{subsec:domain-knowledge}} \addvspace{10pt}
Effective scientific reasoning requires domain knowledge integration beyond basic workflow execution. FlamePilot achieves this through a structured ``skills" system inspired by Claude's Agent Skills \cite{noauthor_claude_2025}, which dynamically augments the agent's capabilities with specialized expertise.

The system organizes knowledge into modular units containing instructions, scripts, and resources that the agent discovers and loads on-demand. We implemented dedicated skills for OpenFOAM and its derivative, DeepFlame. The DeepFlame skill specifically provides the agent with knowledge of how this framework extends from OpenFOAM, including its unique configuration conventions and solver requirements. This modular approach enables operation across different simulation environments using the same foundational toolset, eliminating the need for workflow-specific tools.

Beyond operational knowledge, FlamePilot incorporates research analysis capabilities to ground its decisions in established scientific findings. A dedicated paper analysis skill instructs the agent to process academic PDFs into markdown format and extract critical CFD parameters. This includes geometry and mesh details, model selections, and specific tuning parameters documented in prior studies. By systematically converting raw literature into a structured, queryable format, this skill bridges the gap between published research and practical configuration, allowing the agent to inform its CFD decisions with evidence-based context from the scientific corpus.

\section{Results and discussion\label{sec:results}} \addvspace{10pt}

\subsection{Foundation: FlamePilot's Core Competence in CFD Tasks\label{sec:foambench}} \addvspace{10pt}
The evaluation of FlamePilot's capability for complex OpenFOAM setup and execution utilized the Advanced section of FoamBench from the CFDLLMBench suite \cite{somasekharan_cfdllmbench_2025}. This benchmark comprises 16 expert-crafted OpenFOAM cases that require translating natural language descriptions into intricate actions, including turbulence model selection and mesh generation. 

\begin{figure}[ht!]
\centering
\includegraphics[width=192pt]{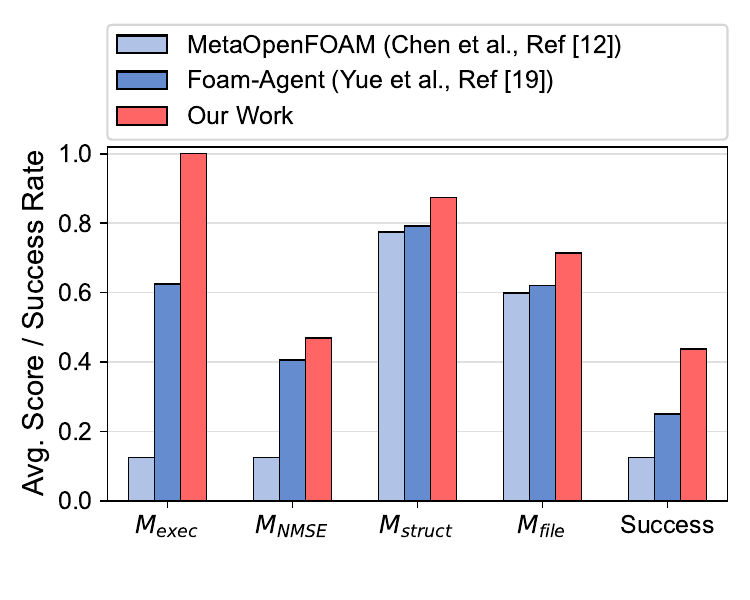}
\caption{\footnotesize Agent performance on FoamBench-Advanced \cite{somasekharan_cfdllmbench_2025} cases. FlamePilot (Our Work) achieves perfect executability ($M_{\text{exec}}$ = 1.0), demonstrating superior operational robustness. Performance metrics for MetaOpenFOAM \cite{chen_metaopenfoam_2025} and Foam-Agent \cite{yue_foam-agent_2025} are based on values reported in \cite{somasekharan_cfdllmbench_2025}.}
\label{foambench}
\end{figure}

FlamePilot was provided only with test queries and access to a local OpenFOAM tutorial directory, receiving no additional human guidance. Its performance was compared against other CFD agents on the same benchmark, as summarized in Fig. \ref{foambench}. Detailed explanation about evaluation metrics could be found in \cite{somasekharan_cfdllmbench_2025}.

FlamePilot achieved a perfect executability score ($M_{\text{exec}}$), successfully producing executable OpenFOAM cases across all 16 tests. This represents a decisive improvement over the previously best-reported benchmark score of 0.625 and directly validates our architectural design; the robustness of the atomic tool integration and single-agent execution model enabled the flexible error correction and rapid iteration required for this performance.

In contrast, while the Success Rate of 0.438 and NMSE-based accuracy ($M_\text{{NMSE}}$) of 0.469 also surpassed the prior state-of-the-art records of 0.250 and 0.406, the absolute values remained low. This discrepancy arises because the $M_\text{{NMSE}}$ metric performs a point-by-point comparison of scalar fields, where even physically plausible solutions can incur high errors due to minor discrepancies in mesh topology, boundary condition placement, or convergence criteria. Our manual, case-by-case analysis verified the physical plausibility of FlamePilot's simulated flow fields identified specific cases where its application of \texttt{snappyHexMesh} for complex geometries resulted in meshes superior to the benchmark's reference.

These results establish FlamePilot's core competence in configuring and executing diverse CFD simulations. While FoamBench mainly focuses on non-reacting flows, this demonstrated proficiency in foundational OpenFOAM operations provides the necessary groundwork for its application to the more complex domain of MILD combustion modeling, as demonstrated in the following section.

\subsection{Case Study: FlamePilot as an Interactive Research Assistant for MILD Combustion Simulation} \label{subsec:case_study} \addvspace{10pt}
This case study demonstrates FlamePilot's transformative role as an interactive research assistant through application to MILD combustion. The investigation showcases FlamePilot's evolution beyond simple automation into a collaborative partner capable of managing complex, iterative workflows grounded in scientific literature.

\begin{figure*}[ht!]
\centering
\vspace{-0.4 in}
\includegraphics[width=400pt]{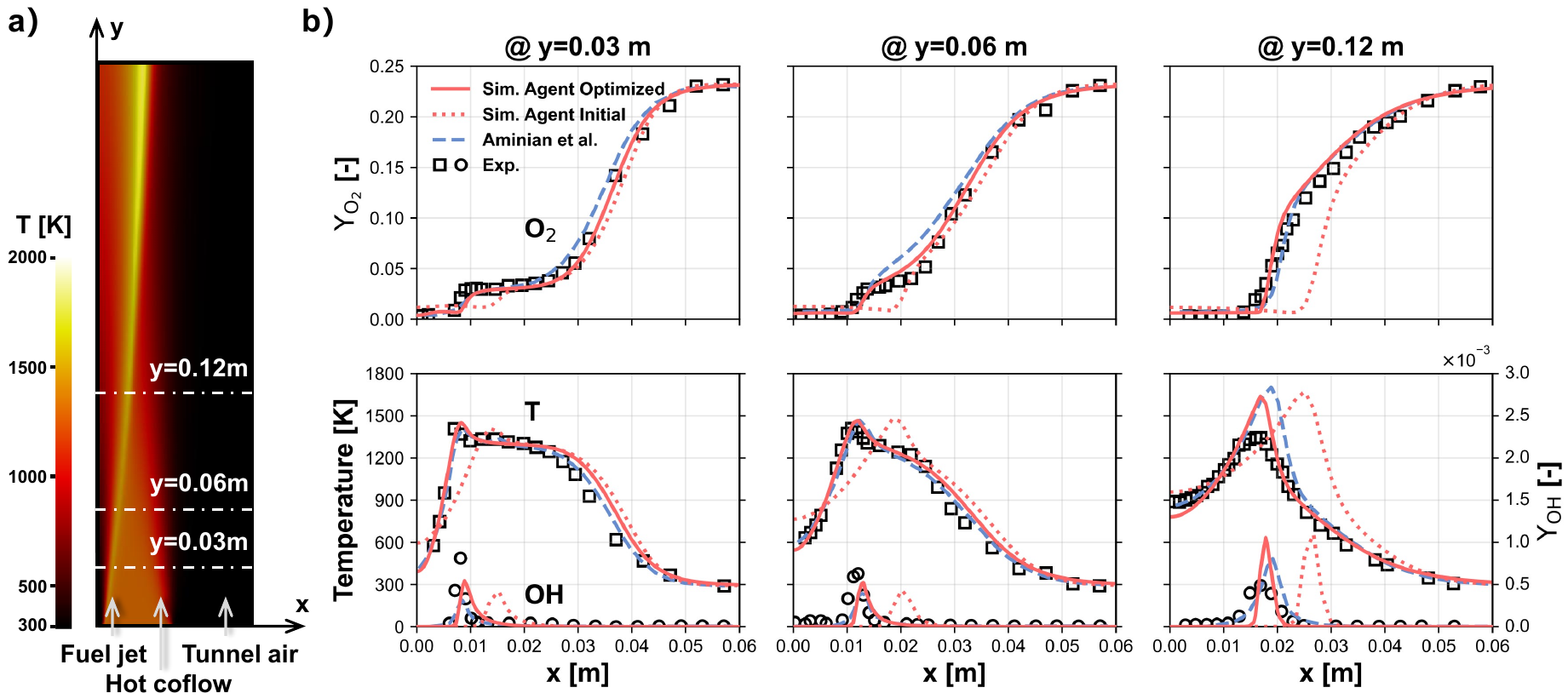}
\vspace{5 pt}
\caption{\footnotesize Comparison of simulation results for the Adelaide JHC MILD burner: (a) Schematic of the 2D computational domain (not to scale); (b) Scalar profiles showing the improvement from initial simulation to final results after literature-guided optimization, demonstrating much better agreement with experimental data \cite{dally_structure_2002}. The complete process from initial setup to final optimization required only a few hours, with simulation runtime being the dominant factor. The optimized results achieve accuracy comparable to established simulation results in the literature \cite{aminian_numerical_2012}.}
\label{mild}
\end{figure*}

The research engagement commenced with a researcher providing a paper on MILD combustion \cite{li_comprehensive_2017} with instructions to reproduce its baseline case. The computation domain is as illustrated in Fig. \ref{mild}a). FlamePilot autonomously processed the PDF to extract critical parameters including geometry, boundary conditions, turbulence-chemistry interaction models, chemical kinetics, and solver specifications, subsequently translating this knowledge into a fully configured and executable DeepFlame case. This process established a direct pipeline from published research to operational simulation, dramatically compressing the timeline from literature review to initial computational results.

When initial simulations showed significant discrepancies with experimental data \cite{dally_structure_2002} (Fig. \ref{mild}b), FlamePilot analyzed provided literature \cite{aminian_numerical_2012,zhao_no_2022,ferrarotti_role_2019} to generate evidence-based improvement proposals. The researcher selected adjustments to the k-epsilon model constant ($C_{1\varepsilon}$) and initiated a parameter study for inlet turbulent kinetic energy, while manually refining mesh quality by modifying two parameters in the pre-configured \texttt{blockMeshDict}. FlamePilot then autonomously executed the complete workflow: implementing model changes, running the parameter study, processing data, and generating visualizations.

The complete process from initial setup to final optimization required only a few hours, with simulation runtime being the dominant factor. The optimized results demonstrate markedly improved agreement with experimental measurements, achieving accuracy comparable to established simulations in the literature \cite{aminian_numerical_2012}. This successful refinement validates FlamePilot's capacity for evidence-based optimization while highlighting its role as a force multiplier in computational research. Throughout this process, the researcher maintained oversight by providing contextual guidance, analyzing results to determine scientific direction, and making critical decisions regarding model adjustments and mesh refinement. FlamePilot managed the substantial burdens of workflow orchestration, postprocessing, and literature-informed hypothesis generation.

\section{Conclusion\label{sec:conclusion}} \addvspace{10pt}
This work presents FlamePilot, an LLM agent that bridges a critical gap in combustion modeling by integrating deep domain knowledge with robust execution for complex tools like OpenFOAM. FlamePilot's architecture demonstrates a practical path toward automated, self-corrective CFD workflows.

Empirical validation confirms FlamePilot's capabilities: it achieved a perfect executability score on the FoamBench-Advanced benchmark, proving its atomic toolset provides the reliability required for research. Furthermore, a MILD combustion case study showed FlamePilot functioning as an interactive research copilot, ingesting literature to configure simulations, proposing evidence-based refinements, and autonomously managing parameter studies to close the loop between published knowledge and validated results.

By employing a single-agent, human-in-the-loop paradigm, FlamePilot ensures transparency and researcher guidance, creating a collaborative partnership that offloads workflow orchestration to accelerate discovery. This work establishes a foundational framework for AI-empowered combustion science. Its design is inherently extendable to diverse CFD codebases and will continuously improve with advances in foundation LLMs, positioning agentic AI as a sustainable force multiplier that upholds scientific rigor and reproducibility.

\acknowledgement{Declaration of competing interest} \addvspace{10pt}
The authors declare no known competing interests.

\acknowledgement{Acknowledgments} \addvspace{10pt}
This work is supported by the National Natural Science Foundation of China (Grant Nos. 92270203, 52276096, and 52441603).

\footnotesize
\baselineskip 9pt

\clearpage
\thispagestyle{empty}
\bibliographystyle{pci}
\bibliography{PCI_LaTeX}

@misc{cantera,
    author = "David G. Goodwin and Harry K. Moffat and Ingmar Schoegl and Raymond L.
              Speth and Bryan W. Weber",
    title = "Cantera: An Object-oriented Software Toolkit for Chemical
             Kinetics, Thermodynamics, and Transport Processes",
    year = 2024,
    note = "Version 3.1.0",
    howpublished = "\url{https://www.cantera.org}",
    doi = {10.5281/zenodo.14455267}
}

@article{jasak_openfoam_2009,
	title = {{OpenFOAM}: {Open} source {CFD} in research and industry},
	volume = {1},
	copyright = {https://www.elsevier.com/tdm/userlicense/1.0/},
	issn = {20926782},
	shorttitle = {{OpenFOAM}},
	url = {https://linkinghub.elsevier.com/retrieve/pii/S2092678216303879},
	doi = {10.2478/IJNAOE-2013-0011},
	language = {en},
	number = {2},
	urldate = {2025-11-11},
	journal = {International Journal of Naval Architecture and Ocean Engineering},
	author = {Jasak, Hrvoje},
	month = dec,
	year = {2009},
	pages = {89--94},
}

@misc{noauthor_claude_2025,
	title = {Claude {Code.} {https://www.anthropic.com/claude-code}},
	url = {https://www.anthropic.com/claude-code},
	year = {2025},
}

@misc{noauthor_codex_2025,
	title = {{Codex.} {https://openai.com/codex/}},
	url = {https://openai.com/codex/},
	year = {2025},
}

@misc{noauthor_gemini_2025,
	title = {Gemini {Code} {CLI.} {https://github.com/google-gemini/gemini-cli}},
	url = {https://github.com/google-gemini/gemini-cli},
	year = {2025},
}

@misc{noauthor_opencode_2025,
	title = {{OpenCode.} {https://github.com/sst/opencode}},
	url = {https://github.com/sst/opencode},
	year = {2025},
}

@article{dally_structure_2002,
	title = {Structure of turbulent non-premixed jet flames in a diluted hot coflow},
	volume = {29},
	copyright = {https://www.elsevier.com/tdm/userlicense/1.0/},
	issn = {15407489},
	url = {https://linkinghub.elsevier.com/retrieve/pii/S1540748902801456},
	doi = {10.1016/S1540-7489(02)80145-6},
	language = {en},
	number = {1},
	urldate = {2025-11-11},
	journal = {Proceedings of the Combustion Institute},
	author = {Dally, B.B. and Karpetis, A.N. and Barlow, R.S.},
	month = jan,
	year = {2002},
	pages = {1147--1154},
}

@misc{huang_biomni_2025,
	title = {Biomni: {A} {General}-{Purpose} {Biomedical} {AI} {Agent}},
	copyright = {http://creativecommons.org/licenses/by/4.0/},
	shorttitle = {Biomni},
	url = {http://biorxiv.org/lookup/doi/10.1101/2025.05.30.656746},
	doi = {10.1101/2025.05.30.656746},
	language = {en},
	urldate = {2025-11-11},
	publisher = {Bioinformatics},
	author = {Huang, Kexin and Zhang, Serena and Wang, Hanchen and Qu, Yuanhao and Lu, Yingzhou and Roohani, Yusuf and Li, Ryan and Qiu, Lin and Li, Gavin and Zhang, Junze and Yin, Di and Marwaha, Shruti and Carter, Jennefer N. and Zhou, Xin and Wheeler, Matthew and Bernstein, Jonathan A. and Wang, Mengdi and He, Peng and Zhou, Jingtian and Snyder, Michael and Cong, Le and Regev, Aviv and Leskovec, Jure},
	month = jun,
	year = {2025},
}

@article{li_comprehensive_2017,
	title = {Comprehensive numerical study of the {Adelaide} {Jet} in {Hot}-{Coflow} burner by means of {RANS} and detailed chemistry},
	volume = {139},
	issn = {03605442},
	url = {https://linkinghub.elsevier.com/retrieve/pii/S0360544217313142},
	doi = {10.1016/j.energy.2017.07.132},
	language = {en},
	urldate = {2025-11-11},
	journal = {Energy},
	author = {Li, Zhiyi and Cuoci, Alberto and Sadiki, Amsini and Parente, Alessandro},
	month = nov,
	year = {2017},
	pages = {555--570},
}

@article{zhao_no_2022,
	title = {{NO} formation mechanism of {CH4}/{NH3} jet flames in hot co-flow under {MILD}-oxy condition: {Effects} of co-flow {CO2} and {H2O}},
	volume = {313},
	issn = {00162361},
	shorttitle = {{NO} formation mechanism of {CH4}/{NH3} jet flames in hot co-flow under {MILD}-oxy condition},
	url = {https://linkinghub.elsevier.com/retrieve/pii/S0016236121028908},
	doi = {10.1016/j.fuel.2021.123030},
	language = {en},
	urldate = {2025-11-11},
	journal = {Fuel},
	author = {Zhao, Zhenghong and Zhang, Tai and Li, Xiaoshan and Zhang, Liqi and Zhang, Zewu and Chen, Yuxiao and Wu, Fan and Luo, Cong and Zheng, Chuguang},
	month = apr,
	year = {2022},
	pages = {123030},
}

@article{aminian_numerical_2012,
	title = {Numerical {Investigation} of a {MILD} {Combustion} {Burner}: {Analysis} of {Mixing} {Field}, {Chemical} {Kinetics} and {Turbulence}-{Chemistry} {Interaction}},
	volume = {88},
	copyright = {http://www.springer.com/tdm},
	issn = {1386-6184, 1573-1987},
	shorttitle = {Numerical {Investigation} of a {MILD} {Combustion} {Burner}},
	url = {http://link.springer.com/10.1007/s10494-012-9386-z},
	doi = {10.1007/s10494-012-9386-z},
	language = {en},
	number = {4},
	urldate = {2025-11-11},
	journal = {Flow Turbulence Combust},
	author = {Aminian, Javad and Galletti, Chiara and Shahhosseini, Shahrokh and Tognotti, Leonardo},
	month = jun,
	year = {2012},
	pages = {597--623},
}

@article{ferrarotti_role_2019,
	title = {On the role of mixing models in the simulation of {MILD} combustion using finite-rate chemistry combustion models},
	volume = {37},
	issn = {15407489},
	url = {https://linkinghub.elsevier.com/retrieve/pii/S1540748918304619},
	doi = {10.1016/j.proci.2018.07.043},
	language = {en},
	number = {4},
	urldate = {2025-11-11},
	journal = {Proceedings of the Combustion Institute},
	author = {Ferrarotti, Marco and Li, Zhiyi and Parente, Alessandro},
	year = {2019},
	pages = {4531--4538},
}

@article{mao_deepflame_2023,
	title = {{DeepFlame}: {A} deep learning empowered open-source platform for reacting flow simulations},
	volume = {291},
	issn = {00104655},
	shorttitle = {{DeepFlame}},
	url = {https://linkinghub.elsevier.com/retrieve/pii/S001046552300187X},
	doi = {10.1016/j.cpc.2023.108842},
	language = {en},
	urldate = {2023-10-16},
	journal = {Computer Physics Communications},
	author = {Mao, Runze and Lin, Minqi and Zhang, Yan and Zhang, Tianhan and Xu, Zhi-Qin John and Chen, Zhi X.},
	month = oct,
	year = {2023},
	pages = {108842},
}

@article{mao_deepflame_2025,
	title = {{DeepFlame} 2.0: {A} new version for fully {GPU}-native machine learning accelerated reacting flow simulations under low-{Mach} conditions},
	volume = {312},
	issn = {00104655},
	shorttitle = {{DeepFlame} 2.0},
	url = {https://linkinghub.elsevier.com/retrieve/pii/S0010465525000980},
	doi = {10.1016/j.cpc.2025.109595},
	language = {en},
	urldate = {2025-05-03},
	journal = {Computer Physics Communications},
	author = {Mao, Runze and Dong, Xinyu and Bai, Xuan and Wu, Ziheng and Dang, Guanlin and Li, Han and Chen, Zhi X.},
	month = jul,
	year = {2025},
	pages = {109595},
}

@article{mao_integrated_2024,
	title = {An integrated framework for accelerating reactive flow simulation using {GPU} and machine learning models},
	volume = {40},
	copyright = {https://www.elsevier.com/tdm/userlicense/1.0/},
	issn = {1540-7489},
	url = {https://linkinghub.elsevier.com/retrieve/pii/S1540748924003201},
	doi = {10.1016/j.proci.2024.105512},
	language = {en},
	number = {1-4},
	urldate = {2025-07-11},
	journal = {Proceedings of the Combustion Institute},
	author = {Mao, Runze and Zhang, Min and Wang, Yingrui and Li, Han and Xu, Jiayang and Dong, Xinyu and Zhang, Yan and Chen, Zhi X.},
	year = {2024},
	note = {Publisher: Elsevier BV},
	pages = {105512},
}

@misc{chai_scimaster_2025,
	title = {{SciMaster}: {Towards} {General}-{Purpose} {Scientific} {AI} {Agents}, {Part} {I}. {X}-{Master} as {Foundation}: {Can} {We} {Lead} on {Humanity}'s {Last} {Exam}?},
	shorttitle = {{SciMaster}},
	url = {http://arxiv.org/abs/2507.05241},
	doi = {10.48550/arXiv.2507.05241},
	urldate = {2025-09-12},
	publisher = {arXiv},
	author = {Chai, Jingyi and Tang, Shuo and Ye, Rui and Du, Yuwen and Zhu, Xinyu and Zhou, Mengcheng and Wang, Yanfeng and E, Weinan and Zhang, Yuzhi and Zhang, Linfeng and Chen, Siheng},
	month = jul,
	year = {2025},
	note = {arXiv:2507.05241 [cs]},
}

@misc{wang_hitchhikers_2025,
	title = {The {Hitchhiker}'s {Guide} to {Autonomous} {Research}: {A} {Survey} of {Scientific} {Agents}},
	copyright = {https://creativecommons.org/licenses/by/4.0/},
	shorttitle = {The {Hitchhiker}'s {Guide} to {Autonomous} {Research}},
	url = {https://www.techrxiv.org/users/951553/articles/1320864-the-hitchhiker-s-guide-to-autonomous-research-a-survey-of-scientific-agents?commit=4d529b07b386cb309b74b765ec8d99fbd6a732c1},
	doi = {10.36227/techrxiv.175459840.02185500/v1},
	urldate = {2025-09-12},
	publisher = {Preprints},
	author = {Wang, Xinming and Xu, Jian and Feng, Aslan H and Chen, Yi and Guo, Haiyang and Zhu, Fei and Shao, Yuanqi and Ren, Minsi and Yi, Hongzhu and Lian, Sheng and Yang, Hongming and Wu, Tailin and Hu, Han and Xiang, Shiming and Zhang, Xu-Yao and Liu, Cheng-Lin},
	month = aug,
	year = {2025},
}

@misc{wei_ai_2025,
	title = {From {AI} for {Science} to {Agentic} {Science}: {A} {Survey} on {Autonomous} {Scientific} {Discovery}},
	shorttitle = {From {AI} for {Science} to {Agentic} {Science}},
	url = {http://arxiv.org/abs/2508.14111},
	doi = {10.48550/arXiv.2508.14111},
	urldate = {2025-09-12},
	publisher = {arXiv},
	author = {Wei, Jiaqi and Yang, Yuejin and Zhang, Xiang and Chen, Yuhan and Zhuang, Xiang and Gao, Zhangyang and Zhou, Dongzhan and Wang, Guangshuai and Gao, Zhiqiang and Cao, Juntai and Qiu, Zijie and He, Xuming and Zhang, Qiang and You, Chenyu and Zheng, Shuangjia and Ding, Ning and Ouyang, Wanli and Dong, Nanqing and Cheng, Yu and Sun, Siqi and Bai, Lei and Zhou, Bowen},
	month = aug,
	year = {2025},
	note = {arXiv:2508.14111 [cs]},
}

@misc{pham_chemgraph_2025,
	title = {{ChemGraph}: {An} {Agentic} {Framework} for {Computational} {Chemistry} {Workflows}},
	shorttitle = {{ChemGraph}},
	url = {http://arxiv.org/abs/2506.06363},
	doi = {10.48550/arXiv.2506.06363},
	urldate = {2025-09-12},
	publisher = {arXiv},
	author = {Pham, Thang D. and Tanikanti, Aditya and Keçeli, Murat},
	month = jun,
	year = {2025},
	note = {arXiv:2506.06363 [physics]},
}

@misc{novikov_alphaevolve_2025,
	title = {{AlphaEvolve}: {A} coding agent for scientific and algorithmic discovery},
	shorttitle = {{AlphaEvolve}},
	url = {http://arxiv.org/abs/2506.13131},
	doi = {10.48550/arXiv.2506.13131},
	urldate = {2025-09-29},
	publisher = {arXiv},
	author = {Novikov, Alexander and Vũ, Ngân and Eisenberger, Marvin and Dupont, Emilien and Huang, Po-Sen and Wagner, Adam Zsolt and Shirobokov, Sergey and Kozlovskii, Borislav and Ruiz, Francisco J. R. and Mehrabian, Abbas and Kumar, M. Pawan and See, Abigail and Chaudhuri, Swarat and Holland, George and Davies, Alex and Nowozin, Sebastian and Kohli, Pushmeet and Balog, Matej},
	month = jun,
	year = {2025},
	note = {arXiv:2506.13131 [cs]},
}

@misc{mitchener_kosmos_2025,
	title = {Kosmos: {An} {AI} {Scientist} for {Autonomous} {Discovery}},
	shorttitle = {Kosmos},
	url = {http://arxiv.org/abs/2511.02824},
	doi = {10.48550/arXiv.2511.02824},
	urldate = {2025-11-06},
	publisher = {arXiv},
	author = {Mitchener, Ludovico and Yiu, Angela and Chang, Benjamin and et al.},
	month = nov,
	year = {2025},
	note = {arXiv:2511.02824 [cs]},
}

@misc{xu_cfdagent_2025,
	title = {{CFDagent}: {A} {Language}-{Guided}, {Zero}-{Shot} {Multi}-{Agent} {System} for {Complex} {Flow} {Simulation}},
	copyright = {arXiv.org perpetual, non-exclusive license},
	shorttitle = {{CFDagent}},
	url = {https://arxiv.org/abs/2507.23693},
	doi = {10.48550/ARXIV.2507.23693},
	urldate = {2025-09-12},
	publisher = {arXiv},
	author = {Xu, Zhaoyue and Wang, Long and Wang, Chunyu and Chen, Yixin and Luo, Qingyong and Yao, Hua-Dong and Wang, Shizhao and He, Guowei},
	year = {2025},
	note = {Version Number: 1},
}

@misc{fan_chatcfd_2025,
	title = {{ChatCFD}: {An} {LLM}-{Driven} {Agent} for {End}-to-{End} {CFD} {Automation} with {Domain}-{Specific} {Structured} {Reasoning}},
	copyright = {Creative Commons Attribution Non Commercial Share Alike 4.0 International},
	shorttitle = {{ChatCFD}},
	url = {https://arxiv.org/abs/2506.02019},
	doi = {10.48550/ARXIV.2506.02019},
	urldate = {2025-09-12},
	publisher = {arXiv},
	author = {Fan, E and Hu, Kang and Wu, Zhuowen and Ge, Jiangyang and Miao, Jiawei and Zhang, Yuzhi and Sun, He and Wang, Weizong and Zhang, Tianhan},
	year = {2025},
	note = {Version Number: 2},
}

@article{pandey_openfoamgpt_2025,
	title = {{OpenFOAMGPT}: {A} retrieval-augmented large language model ({LLM}) agent for {OpenFOAM}-based computational fluid dynamics},
	volume = {37},
	issn = {1070-6631, 1089-7666},
	shorttitle = {{OpenFOAMGPT}},
	url = {https://pubs.aip.org/pof/article/37/3/035120/3338372/OpenFOAMGPT-A-retrieval-augmented-large-language},
	doi = {10.1063/5.0257555},
	language = {en},
	number = {3},
	urldate = {2025-09-12},
	journal = {Physics of Fluids},
	author = {Pandey, Sandeep and Xu, Ran and Wang, Wenkang and Chu, Xu},
	month = mar,
	year = {2025},
	pages = {035120},
}

@misc{chen_optmetaopenfoam_2025,
	title = {{OptMetaOpenFOAM}: {Large} {Language} {Model} {Driven} {Chain} of {Thought} for {Sensitivity} {Analysis} and {Parameter} {Optimization} based on {CFD}},
	copyright = {arXiv.org perpetual, non-exclusive license},
	shorttitle = {{OptMetaOpenFOAM}},
	url = {https://arxiv.org/abs/2503.01273},
	doi = {10.48550/ARXIV.2503.01273},
	urldate = {2025-09-12},
	publisher = {arXiv},
	author = {Chen, Yuxuan and Zhang, Long and Zhu, Xu and Zhou, Hua and Ren, Zhuyin},
	year = {2025},
	note = {Version Number: 1},
}

@misc{yue_foam-agent_2025,
	title = {Foam-{Agent}: {Towards} {Automated} {Intelligent} {CFD} {Workflows}},
	copyright = {Creative Commons Attribution Non Commercial No Derivatives 4.0 International},
	shorttitle = {Foam-{Agent}},
	url = {https://arxiv.org/abs/2505.04997},
	doi = {10.48550/ARXIV.2505.04997},
	urldate = {2025-09-12},
	publisher = {arXiv},
	author = {Yue, Ling and Somasekharan, Nithin and Cao, Yadi and Pan, Shaowu},
	year = {2025},
	note = {Version Number: 1},
}

@misc{chen_metaopenfoam_2025,
	title = {{MetaOpenFOAM} 2.0: {Large} {Language} {Model} {Driven} {Chain} of {Thought} for {Automating} {CFD} {Simulation} and {Post}-{Processing}},
	copyright = {arXiv.org perpetual, non-exclusive license},
	shorttitle = {{MetaOpenFOAM} 2.0},
	url = {https://arxiv.org/abs/2502.00498},
	doi = {10.48550/ARXIV.2502.00498},
	urldate = {2025-09-12},
	publisher = {arXiv},
	author = {Chen, Yuxuan and Zhu, Xu and Zhou, Hua and Ren, Zhuyin},
	year = {2025},
	note = {Version Number: 1},
}

@misc{chen_metaopenfoam_2024,
	title = {{MetaOpenFOAM}: an {LLM}-based multi-agent framework for {CFD}},
	copyright = {arXiv.org perpetual, non-exclusive license},
	shorttitle = {{MetaOpenFOAM}},
	url = {https://arxiv.org/abs/2407.21320},
	doi = {10.48550/ARXIV.2407.21320},
	urldate = {2025-09-12},
	publisher = {arXiv},
	author = {Chen, Yuxuan and Zhu, Xu and Zhou, Hua and Ren, Zhuyin},
	year = {2024},
	note = {Version Number: 2},
}

@misc{xu_llm_2024,
	title = {{LLM} {Agent} for {Fire} {Dynamics} {Simulations}},
	copyright = {arXiv.org perpetual, non-exclusive license},
	url = {https://arxiv.org/abs/2412.17146},
	doi = {10.48550/ARXIV.2412.17146},
	urldate = {2025-09-12},
	publisher = {arXiv},
	author = {Xu, Leidong and Mohaddes, Danyal and Wang, Yi},
	year = {2024},
	note = {Version Number: 1},
}

@article{dong_fine-tuning_2025,
	title = {Fine-tuning a large language model for automating computational fluid dynamics simulations},
	volume = {15},
	issn = {20950349},
	url = {https://linkinghub.elsevier.com/retrieve/pii/S2095034925000261},
	doi = {10.1016/j.taml.2025.100594},
	language = {en},
	number = {3},
	urldate = {2025-09-12},
	journal = {Theoretical and Applied Mechanics Letters},
	author = {Dong, Zhehao and Lu, Zhen and Yang, Yue},
	month = may,
	year = {2025},
	pages = {100594},
}

@misc{feng_openfoamgpt_2025,
	title = {{OpenFOAMGPT} 2.0: end-to-end, trustworthy automation for computational fluid dynamics},
	shorttitle = {{OpenFOAMGPT} 2.0},
	url = {http://arxiv.org/abs/2504.19338},
	doi = {10.48550/arXiv.2504.19338},
	urldate = {2025-09-12},
	publisher = {arXiv},
	author = {Feng, Jingsen and Xu, Ran and Chu, Xu},
	month = apr,
	year = {2025},
	note = {arXiv:2504.19338 [physics]},
}

@misc{somasekharan_cfdllmbench_2025,
	title = {{CFDLLMBench}: {A} {Benchmark} {Suite} for {Evaluating} {Large} {Language} {Models} in {Computational} {Fluid} {Dynamics}},
	shorttitle = {{CFDLLMBench}},
	url = {http://arxiv.org/abs/2509.20374},
	doi = {10.48550/arXiv.2509.20374},
	urldate = {2025-11-06},
	publisher = {arXiv},
	author = {Somasekharan, Nithin and Yue, Ling and Cao, Yadi and Li, Weichao and Emami, Patrick and Bhargav, Pochinapeddi Sai and Acharya, Anurag and Xie, Xingyu and Pan, Shaowu},
	month = oct,
	year = {2025},
	note = {arXiv:2509.20374 [cs]},
}


\newpage

\small
\baselineskip 10pt


\end{document}